\documentclass[10pt,twocolumn,letterpaper]{article}

\usepackage[pagenumbers]{cvpr} % To force page numbers, e.g. for an arXiv version

\definecolor{cvprblue}{rgb}{0.21,0.49,0.74}
\usepackage[pagebackref,breaklinks,colorlinks,allcolors=cvprblue]{hyperref}
\usepackage{float}
\usepackage{stfloats}

\def\paperID{11682} % *** Enter the Paper ID here
\def\confName{CVPR}
\def\confYear{2026}

\title{ASTRA: Enhancing Multi-Subject Generation with Retrieval-Augmented Pose Guidance and Disentangled Position Embedding}

\author{
Tianze Xia$^{1*}$	 \quad
Zijian Ning$^{1*}$ \quad
Zonglin Zhao$^{1*}$ \quad
Mingjia Wang$^{1}$  \\
\\
$^{1}$Huazhong University of Science and Technology \\
}

\begin{document}

\twocolumn[{%
    \renewcommand\twocolumn[1][]{#1}%
    \maketitle
    \vspace{-1.5em}  % 调整标题与图像间距
    \begin{center}
        \includegraphics[width=1\textwidth]{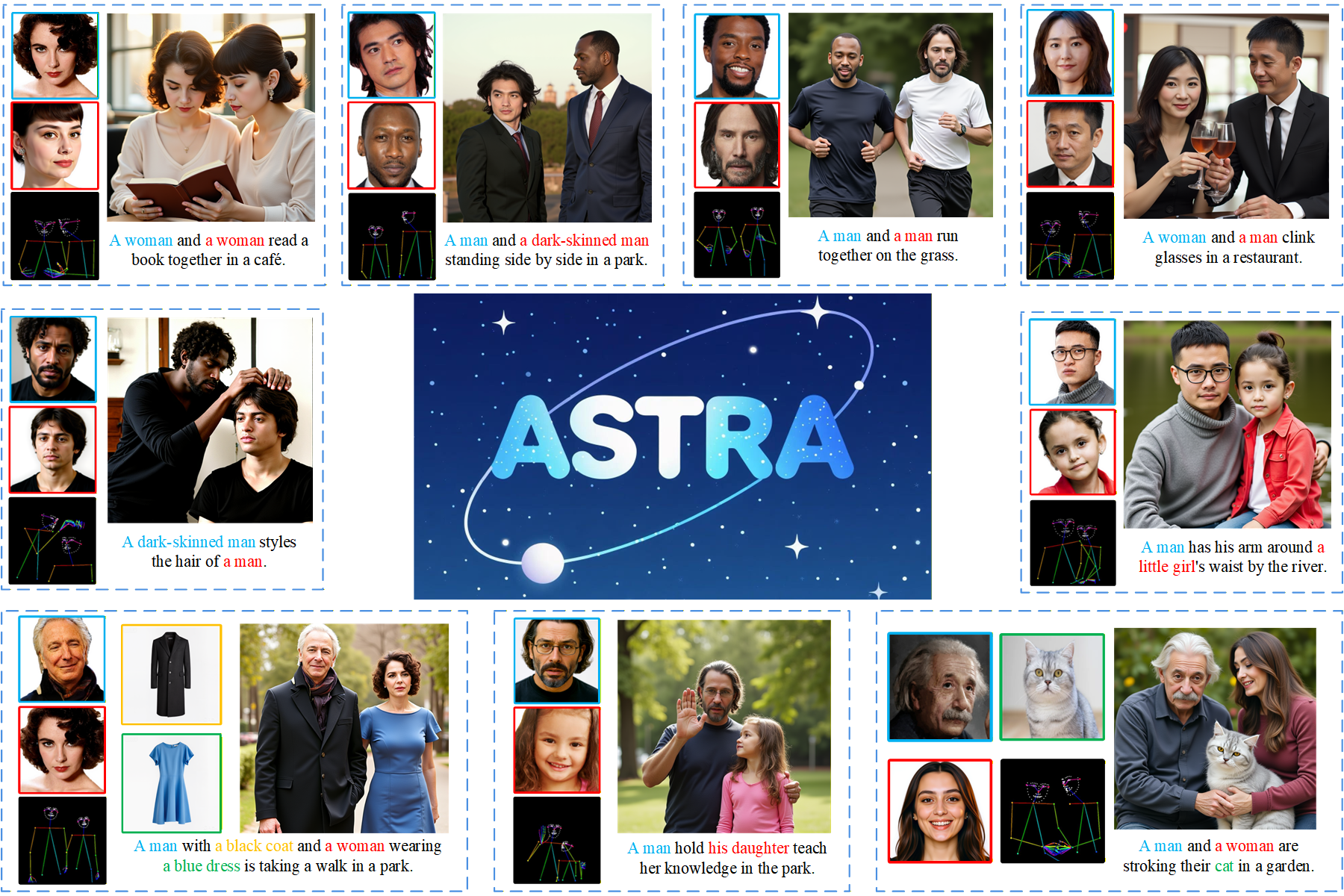}
        \vspace{-0.5em}
        \captionof{figure}{
            Representative outputs showcase the capabilities of \textbf{ASTRA} in multi-subject generation,
            demonstrating both high subject fidelity and precise pose control.In each example, the pose maps presented are retrieved from our meticulously constructed database based on the prompts, and these pose maps provide clear structural guidance for the image synthesis process.
        }
        \label{teaser}
        \vspace{1em}
    \end{center}
}]

\begin{abstract}
Subject-driven image generation has shown great success in creating personalized content, but its capabilities are largely confined to single subjects in common poses. Current approaches face a fundamental conflict when handling multiple subjects with complex, distinct actions: preserving individual identities while enforcing precise pose structures. This challenge often leads to identity fusion and pose distortion, as appearance and structure signals become entangled within the model's architecture. To resolve this conflict, we introduce \textbf{ASTRA}(\textbf{A}daptive \textbf{S}ynthesis through \textbf{T}argeted \textbf{R}etrieval \textbf{A}ugmentation), a novel framework that architecturally disentangles subject appearance from pose structure within a unified Diffusion Transformer. ASTRA achieves this through a dual-pronged strategy. It first employs a Retrieval-Augmented Pose (RAG-Pose) pipeline to provide a clean, explicit structural prior from a curated database. Then, its core generative model learns to process these dual visual conditions using our Enhanced Universal Rotary Position Embedding (EURoPE), an asymmetric encoding mechanism that decouples identity tokens from spatial locations while binding pose tokens to the canvas. Concurrently, a Disentangled Semantic Modulation (DSM) adapter offloads the identity preservation task into the text conditioning stream. Extensive experiments demonstrate that our integrated approach achieves superior disentanglement. On our designed COCO-based complex pose benchmark, ASTRA achieves a new state-of-the-art in pose adherence, while maintaining high identity fidelity and text alignment in DreamBench.
\end{abstract}
\section{Introduction}
\label{sec:intro}

Recent text-to-image models \cite{sohl2015deep, ho2020denoising, esser2024scaling, flux2024} have enabled powerful personalized image generation. While early methods achieved remarkable single-subject control \cite{gal2022image, ruiz2023dreambooth, ye2023ip}, generating coherent scenes with multiple subjects in complex interactions remains a major challenge. This task exposes a core conflict in existing paradigms: preserving multi-subject identity while ensuring complex pose fidelity.

This difficulty stems from two fundamental bottlenecks. First, a data bottleneck: High-quality training data for multi-subject interactions is scarce and impractical to collect, limiting model generalization. Second, an architectural bottleneck: State-of-the-art methods \cite{tan2024ominicontrol, chen2025unireal, wu2025less, mou2025dreamo} that inject identity via attention mechanisms in Diffusion Transformers (DiTs) \cite{peebles2023scalable} often suffer from attention confusion. This leads to artifacts like ``identity fusion'' and ``pose distortion'' , highlighting the need for a new approach to decouple identity and pose control.

To address these bottlenecks, we draw inspiration from Retrieval-Augmented Generation (RAG) \cite{lewis2020retrieval, gao2023retrieval} in LLMs and propose a ``retrieve-then-generate'' framework. Instead of generating complex poses from scratch a process hindered by data scarcity and prone to hallucination \cite{ji2023survey}—we first retrieve a suitable pose structure from a pre-built ``pose knowledge database.'' This strategy circumvents the data problem and decouples the entangled appearance-structure signals at the input level. We introduce ASTRA, a framework that uses dual visual inputs: (1) a Retrieval-Augmented Pose Pipeline provides explicit structural priors from a high-quality text-pose database, and (2) a model framework with our proposed 	\textbf{E}nhanced 	\textbf{U}niversal 	\textbf{Ro}tary 	\textbf{P}osition 	\textbf{E}mbedding 
(\textbf{EURoPE}) and \textbf{D}isentangled \textbf{S}emantic \textbf{M}odulation(\textbf{DSM}) achieves precise control over subject identities and poses.

This decoupling of appearance and structure provides fine-grained, assignable control that current methods lack. While tools like ControlNet \cite{zhang2023adding} master structural guidance, they cannot customize subject identity. Conversely, IP-Adapter \cite{ye2023ip} preserves a single subject's identity but struggles to assign distinct poses to multiple individuals. Our framework ASTRA is the first to allow users to explicitly assign a unique, complex pose to multiple customized subjects simultaneously(as shown in Figure~\ref{teaser}). This capability is crucial for complex storytelling and personalized content creation, overcoming the limitations of text-only and single-subject-only descriptions for such scenes \cite{brooks2023instructpix2pix}.

Our contributions are summarized as follows:
\begin{itemize}
    \item We introduce a RAG-inspired paradigm that addresses the data scarcity of complex actions by retrieving structural priors from a high-quality, curated text-pose database.
    \item We develop ASTRA, a generative architecture with an asymmetric encoding mechanism and Disentangled Semantic Modulation, enabling precise and assignable control over multiple subjects' identities and poses.
    \item On established benchmarks including DreamBench\cite{ruiz2023dreambooth} and multi-subject driven generation benchmarks, our ASTRA achieves highest DINO and CLIP-I scores. Furthermore, in our proposed Controlled Composition on COCO evaluation, our method attains the highest Object Keypoint Similarity (OKS) and CLIP-I score, while securing the second-highest CLIP-T score. These demonstrate its strong subject fidelity, text and pose control ability, also showcasing its capability to deliver state-of-the-art (SOTA) results.
\end{itemize}
\section{Related Work}
\label{sec:related work}
\paragraph{Subject-Driven and Controllable Generation.}
The pursuit of personalized visual content has led to significant progress in subject-driven generation. This field is broadly divided into two main paradigms: Fine-tuning-based methods, including Textual Inversion~\cite{gal2022image} and DreamBooth~\cite{ruiz2023dreambooth}, which perform efficient per-subject tuning to embed novel concepts; and Tuning-free methods, such as IP-Adapter~\cite{ye2023ip}, which utilize a dedicated image encoder to inject subject features. Concurrently, a separate line of work on controllable generation, championed by ControlNet~\cite{zhang2023adding}, has achieved precise spatial control by conditioning on structural maps such as human poses.However, extension of these methods to multiple subjects introduces significant challenges, such as identity bleeding and attribute confusion. To overcome these complexities, recent works have proposed several novel strategies. For instance, MS-Diffusion~\cite{wang2024ms} introduces a layout-guided framework, using a "Grounding Resampler" and masked cross-attention to mitigate subject conflicts. UNO~\cite{wu2025less} proposes a model-data co-evolution paradigm, leveraging a progressive pipeline to synthesize high-consistency multi-subject training data. OmniGen2~\cite{wu2025omnigen2} addresses this as part of a unified model, using a decoupled architecture and low-level VAE features to preserve fine-grained identity during its in-context generation tasks. While these methods advance multi-subject identity preservation, a critical challenge lies at the intersection of these domains: \textbf{Existing methods struggle to simultaneously preserve multiple, specific subject identities while enforcing complex, structured poses}. This issue is particularly acute in modern Diffusion Transformers (DiTs)~\cite{peebles2023scalable}, highlighting the need for a new architecture that can fundamentally decouple appearance from structure.
\paragraph{Retrieval-Augmented Generation (RAG).}
Inspired by its success in Large Language Models (LLMs), Retrieval-Augmented Generation (RAG) \cite{lewis2020retrieval,gao2023retrieval} is an emerging paradigm in visual synthesis. Although prior work has explored retrieval to improve rare concept generation or style transfer~\cite{sheynin2022knn,blattmann2022retrieval,hu2024instruct,shalev2025imagerag}, our work introduces a novel application of RAG philosophy. We leverage it to address the structural data scarcity bottleneck in complex action synthesis. By building and retrieving from a curated “text-pose" database, we provide an explicit, clean structural prior to the model. This strategic, pre-generation retrieval not only bypasses the difficulty of synthesizing complex poses from scratch but is also the cornerstone of our framework's ability to cleanly separate structural information (the pose) from appearance information (the subject's identity), thereby preventing the feature entanglement that plagues existing methods.

\begin{figure*}[t]
\centering
\includegraphics[width=\textwidth]{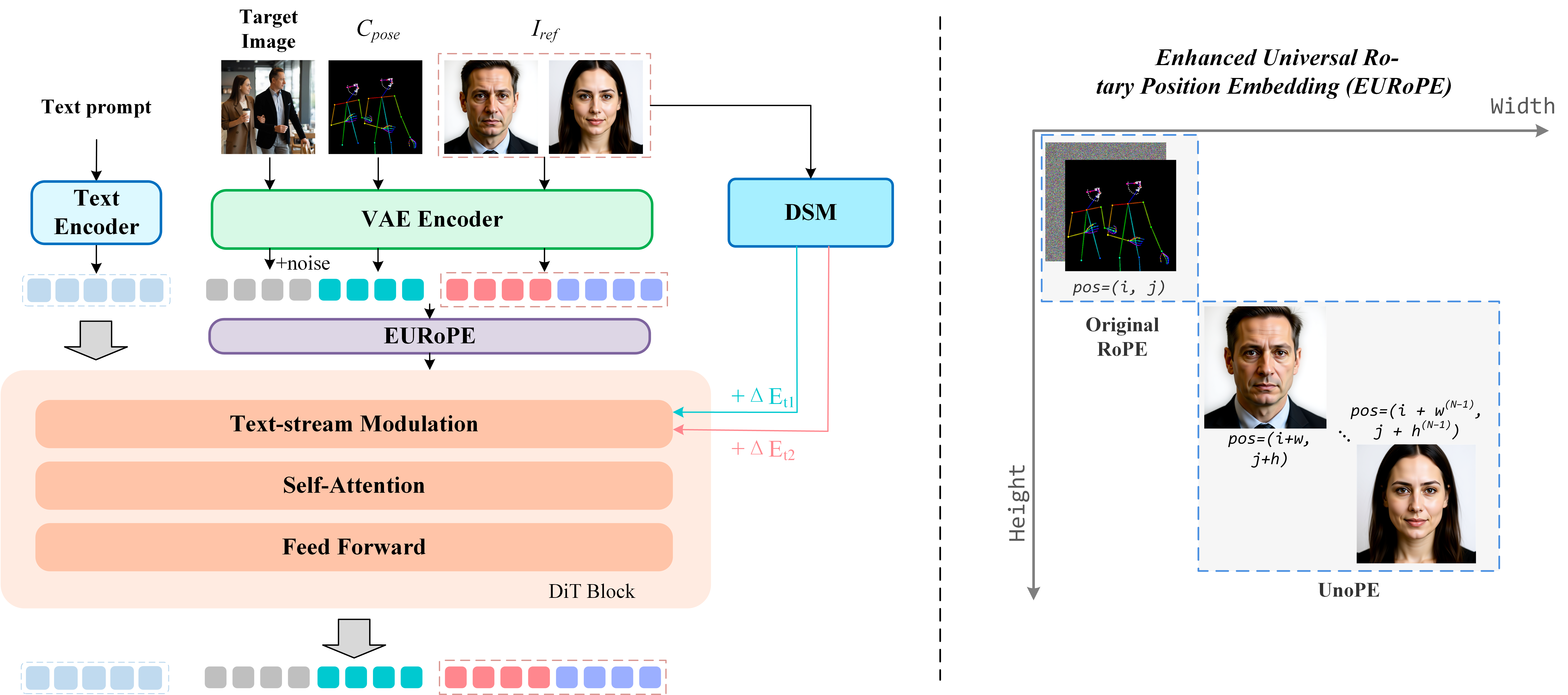} % Reduce the figure size so that it is slightly narrower than the column.
\caption{The ASTRA framework for multi-subject pose-controllable generation. The left panel shows the overall architecture, which uniquely conditions a DiT on reference identities ($I_{	\text{ref}}$), a text prompt, and an explicit pose structure ($C_{	\text{pose}}$). Its effectiveness stems from our Enhanced Universal Rotary Position Embedding (EURoPE), detailed on the right. EURoPE employs an asymmetric strategy: it decouples identity from spatial location using universal rotary position embedding (UnoPE) for $I_{	\text{ref}}$ tokens, while enforcing strict structural constraints by binding $C_{\text{pose}}$ tokens to the latent canvas with native RoPE. Concurrently, Disentangled Semantic Modulation (DSM) offloads identity preservation by modulating text embeddings with visual features. }
\label{fig2}
\end{figure*}
\section{Method}
\label{sec:method}
Modern diffusion transformers (DiT) excel at text-to-image generation, but struggle when simultaneously conditioned on multiple subject identities and precise structural guidance (pose maps). Shared self-attention treats all conditional inputs uniformly, causing feature entanglement—either compromising pose to preserve identity or blending identities incorrectly. ASTRA addresses this by separating structural and appearance signals at both the architectural and representational levels (Figure~\ref{fig2}).
\subsection{Pose Retrieval-Augmented Framework}
\label{sec:retrieval_framework}
\begin{figure}[t]
\centering
\includegraphics[width=0.48\textwidth]{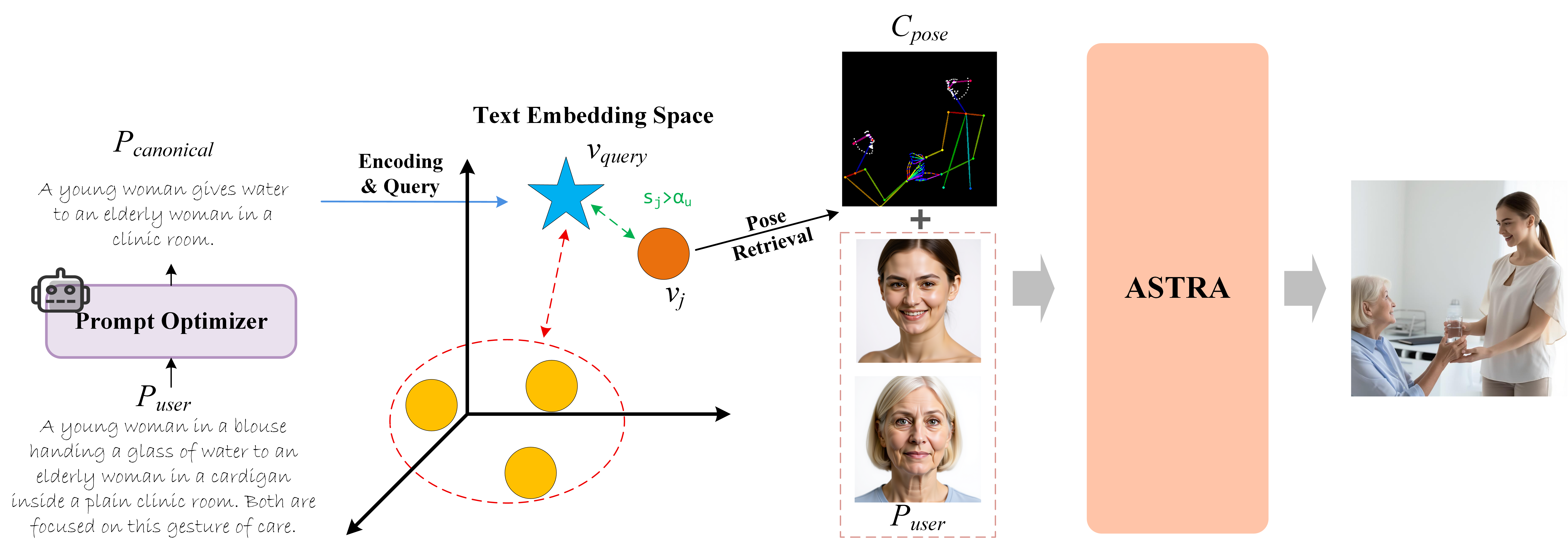} % Reduce the figure size so that it is slightly narrower than the column.
\caption{The inference pipeline of ASTRA. A user prompt is first refined into a canonical query by a VLM-based optimizer to isolate pose-defining semantics. This query facilitates a vector search on our curated database to find an optimal pose map ($C_{	\text{pose}}$). The retrieved pose, acting as explicit structural guidance, is then passed to the ASTRA model along with the subject identities ($I_{	\text{ref}}$) and original text to synthesize the final image.}
\label{fig3}
\end{figure}

The paucity of high-quality, subject-consistent datasets has long presented a formidable obstacle for subject-driven generation. This challenge is particularly acute in the domain of pose estimation and synthesis, where no large-scale, high-fidelity dataset of pose maps currently exists to support robust model training and inference. Compounding this issue, the field of vision-language retrieval—particularly for image-based Retrieval-Augmented Generation (RAG) systems—lacks standardized, end-to-end methodologies for constructing high-quality databases and implementing efficient retrieval mechanisms. Our proposed framework directly addresses these critical gaps by delivering a comprehensive solution that includes: (1) a large-scale, curated pose database with diverse and anatomically accurate 2D skeleton maps; (2) a simple but efficient curation pipeline ensuring semantic fidelity and data quality; and (3) an efficient, lightweight retrieval system that seamlessly maps ambiguous textual descriptions to the most relevant pose representations.The entire process is visualized in Figure~\ref{fig3}.
\paragraph{Systematic and Diverse Prompt Engineering.}
Our process began with a foundation of over 300 distinct human actions, covering a wide spectrum of complexity from simple gestures to intricate multi-person interactions. To ensure semantic richness and diversity, we then algorithmically generated approximately 30 textual variations for each action. This was achieved by systematically manipulating a set of descriptive axes, including subject attributes (e.g., age, body type), action specifics (e.g., "stretching" vs. "stretching lazily"), environmental context, and camera perspective. This systematic generation resulted in a comprehensive set of over 9,000 unique prompts, forming the basis for our image generation phase.
% In your preamble, ensure you have the amsmath package for equations:
% \usepackage{amsmath}
\paragraph{High-Fidelity Generation and Curation.}
We utilized the state-of-the-art text-to-image model, FLUX.1-pro\cite{flux2024}, to generate an initial image pool from our prompt set. To distill this raw collection into a high-quality dataset, we developed a scalable, VLM-driven curation pipeline.
The core of our pipeline is an automated semantic verification stage using the multi-modal capabilities of GPT-4o\cite{hurst2024gpt}. This stage systematically measures the alignment between a generated image and its prompt by evaluating three key dimensions: subject consistency ($s_1$), interaction logic ($s_2$), and detail fidelity ($s_3$). An overall semantic alignment score $S$ is then computed as a weighted sum:
\begin{equation}
S = \omega_1 s_1 + \omega_2 s_2 + \omega_3 s_3
\end{equation}
These components are precisely calibrated. Scores $s_i$ are generated on a continuous scale $s \in [0, 1]$ by instructing the VLM, via a structured prompt, to provide a final numerical rating after its qualitative analysis. The weights ($\omega_1, \omega_2, \omega_3$) were optimized on a held-out validation set against human preference scores, using regression to find the best fit. This resulted in a hierarchy that prioritizes subject and interaction correctness ($\omega_1 > \omega_2 > \omega_3$). The final acceptance threshold $	\theta$ was set to maximize the F1-score on the same validation set.
To validate the effectiveness of this automated pipeline, we conducted a limited human evaluation on the VLM-filtered dataset, checking for subtle errors in pose, composition, and aesthetics. The results showed that fewer than 3\% of the filtered images were flagged as low-quality, confirming the VLM filter's effectiveness in producing high-quality training data.

\paragraph{Pose Extraction and Data Indexing.}
For each image that passed the curation stages, we used OpenPose\cite{cao2017realtime} to extract 2D pose skeletons and added a two-stage pose quality check, with only 3.2\% of pose maps flagged as low-quality, validating the effectiveness of our pipeline.For any prompt whose corresponding pose map was rejected at this stage, the entire generation and two-stage curation cycle was re-initiated until a valid pose representation was successfully produced, ensuring complete and high-quality coverage for our prompt set. Finally, for each verified pose map, the associated text prompt was encoded into a 384-dimensional dense vector using the lightweight all-MiniLM-L6-v2\cite{reimers-2019-sentence-bert} sentence-transformer model. These vectors were then L2-normalized\cite{cortes2012l2} and stored in an efficient vector index, linking each textual representation to its corresponding high-fidelity pose map.
\subsubsection{Retrieval at Inference Time.}
During inference, a  retriever is employed to source an optimal structural prior from our pre-compiled database for any given user prompt. To maximize retrieval precision, we introduce a prompt normalization stage. Raw user prompts often entangle core pose-defining semantics with non-structural information. Therefore, we utilize a lightweight LLM, Qwen2.5-1.5B-Instruct\cite{qwen2.5}, to parse user input $P_{\text{user}}$ and distill it into a canonical query $P_{\text{canonical}}$. 
The objective of this normalization is to generate a query that aligns with the multi-axial descriptive schema used for our database construction. Specifically, the LLM is instructed to preserve semantically rich descriptors pertinent to the pose---such as subject-action nuance, demographic attributes, dynamic features, and compositional directives.
This normalized query, $P_{	\text{canonical}}$, is then encoded into a fixed-size semantic vector, $\mathbf{v}_{	\text{query}}$, using an all-MiniLM-L6-v2 sentence-transformer followed by mean pooling and L2 normalization. This vector facilitates a Maximum Inner Product Search  (MIPS)\cite{shrivastava2014asymmetric} against the indexed database vectors ($\mathbf{v}_j$), equivalent to a cosine similarity search, to identify the entry ($j^*$) with the highest score:
\begin{equation}
j^* = \arg\max_j (\mathbf{v}_{	\text{query}} \cdot \mathbf{v}_j)
\end{equation}
To ensure relevance, we apply a confidence gating mechanism: the corresponding pose map, $C_{	\text{pose}}$, is selected only if the score $s_{j^*}$ exceeds a predefined threshold ($\alpha_u = 0.55$). 	This value was empirically determined on a validation set to optimally balance precision  and recall. If no score meets this threshold---a scenario that typically arises for out-of-distribution prompts---the retrieval step is bypassed. In this case, the model defaults to relying solely on the text prompt for structural guidance. This two-step process---semantic normalization followed by vector search---provides the generation model with an explicit structural condition ($C_{	\text{pose}}$), establishing a robust foundation for high-fidelity synthesis.

\subsubsection{Training Data Construction Pipeline}
To construct training data pairs $\{\text{prompt}, \text{ref\_imgs},\text{pose}\}$, we design a pipeline based on the curated pose database. Target images match specific pose structures. Descriptive prompts are extracted using Florence-2\cite{xiao2023florence}'s captioning module, while its object detection module identifies subject labels and bounding boxes from target images. These are processed by SAM2\cite{ravi2024sam2} to segment individual subjects, which are then re-edited by Flux.1 Kontext to generate diverse reference images. Pose maps are extracted from the target images using OpenPose. Finally, reference images are filtered by CLIP-I scores to ensure alignment with prompts. The resulting data pairs provide a robust foundation for training, integrating textual, structural, and visual information.
\subsection{ASTRA: The Generation Model Framework}
ASTRA trains a DiT-based generator to synthesize images faithful to multiple identities and precise poses. All training data originates from the curated retrieval-augmented database.
\paragraph{Unified Multi-Modal Conditioning and Training.}
In subject-driven generation tasks, a key challenge lies in integrating diverse input modalities—text prompts, noisy latents, and multimodal visual references—while maintaining training stability and generation quality. Traditional methods, such as DreamBooth \cite{ruiz2023dreambooth} and LoRA \cite{hu2022lora}, typically rely on cross-attention mechanisms to inject identity features. However, when scaled to multi-subject scenarios, these approaches encounter issues such as computational complexity and feature entanglement, where each additional subject requires an independent attention path. Recent methods propose two main improvements: converting visual features into token sequences for cross-attention interactions with text \cite{li2024photomaker}, and directly concatenating all input conditions into a unified sequence processed by the model's self-attention mechanism \cite{wu2025less}. We adopt the latter approach for its efficiency, though directly introducing noise-free reference tokens into a model pretrained on noisy data can lead to instability. While previous methods address this through multi-stage training, ASTRA resolves this by explicitly defining the roles of visual conditions at the architectural level, leveraging an asymmetric positional encoding mechanism to enable direct and efficient single-stage training.
We formalize the input sequence construction process as follows. First, we tokenize the reference images $I_{\text{ref}}$ and the retrieved pose map $C_{\text{pose}}$:
\begin{equation}
t_{\text{ref}_i} = E_{\text{img}}(I_{\text{ref}_i}), \quad t_{\text{pose}} = E_{\text{pose}}(C_{\text{pose}})
\end{equation}
These tokens are then assembled into an ordered visual conditioning sequence $C_{\text{visual}}$, expressed as:
\begin{equation}
C_{\text{visual}} = \langle t_{\text{ref}_1}, \ldots, t_{\text{ref}_{N-1}}, t_{\text{pose}} \rangle
\end{equation}
Finally, we concatenate the text condition tokens $c$, visual conditioning sequence $C_{\text{visual}}$, and noisy latent $z_t$ into the complete input stream $z_{\text{in}}$:
\begin{equation}
z_{\text{in}} = c \oplus C_{\text{visual}} \oplus z_t
\end{equation}
\paragraph{Enhanced Universal Rotary Position Embedding (EURoPE).}
As shown in the right panel of Figure~\ref{fig2}, EURoPE is an asymmetric positional encoding mechanism that resolves the conflict between preserving subject appearance and enforcing precise pose geometry. It applies different encoding strategies to identity reference tokens and pose map tokens within the DiT attention mechanism.

For \textbf{reference images}, the goal is to extract identity features while avoiding fixed spatial layout signals that may conflict with text or pose constraints. We adopt Universal Rotary Position Embedding (UnoPE)~\cite{wu2025less}, which sequentially re-indexes reference tokens:
\begin{equation}
(i', j') = (i + w_{N-1}, j + h_{N-1})
\end{equation}
where $w,h$ denote cumulative dimensions of preceding references. This enforces layout control via text/pose while keeping subject identity independent of original image arrangement.

For \textbf{pose maps}, strict spatial binding is desirable. We apply native RoPE~\cite{su2024roformer} with positional indices $(i,j)$ aligned exactly to corresponding patches in the noisy latent $z_t$, forcing strong attention at skeleton locations. Because the pose maps are clean skeletons on neutral backgrounds, positional binding enforces geometry without introducing unwanted details—turning a common positional conflict into a precise structural constraint.
\begin{figure*}[t]
\centering
\includegraphics[width=\textwidth]{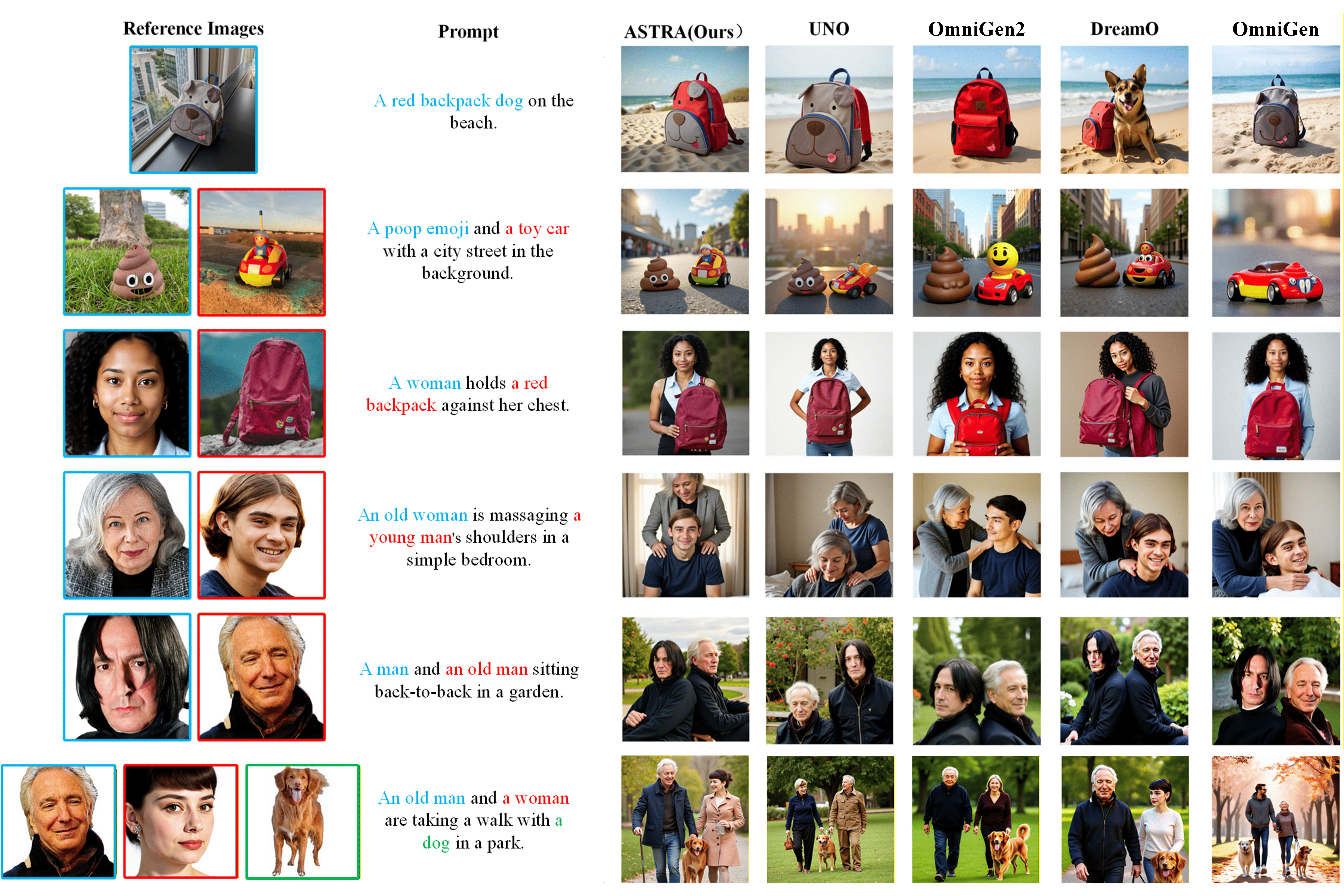}
 \caption{Qualitative comparison with different methods on single and multi subject driven generation}
 \label{figure4}
\end{figure*}
\paragraph{Disentangled Semantic Modulation.}
While the EURoPE encoding mechanism effectively separates structural and appearance signals at the input level, representational entanglement within the self-attention layers of DiT remains a critical challenge. The model must simultaneously process global text semantics, preserve multiple high-fidelity identities, and enforce precise structural constraints, all within a unified feature space. To address this, we propose the Disentangled Semantic Modulation (DSM) stream, a novel module designed to offload identity preservation from the primary denoising pathway. By recontextualizing visual identity as a learnable modulation of the textual condition, DSM explicitly encodes identity information into the semantic instructions guiding the generation process.
The core premise behind DSM is that a subject’s unique identity can be conceptualized as a distinct point in a semantic subspace of the model’s multimodal feature space. DSM distills this identity, extracted from a reference image, into a semantic offset vector $\Delta E_t$. This offset modifies the text embeddings $E_t$ by introducing subject-specific biases. Formally, the DSM module $\Phi$ maps the visual features $F_v$ and text embeddings $E_t$ as:
\begin{equation}
\Delta E_t = \Phi(F_v, E_t; \theta_\Phi),
\end{equation}
where $\theta_\Phi$ represents the trainable parameters of $\Phi$. The modulated embeddings are obtained as:
\begin{equation}
E'_t = E_t + \Delta E_t.
\end{equation}
This mechanism embeds the subject's identity directly into the textual instructions, eliminating the need for the primary DiT blocks to learn implicit correlations between reference tokens and image outputs. Instead, identity is seamlessly integrated into the semantic control signals.
The DSM module is implemented as a lightweight, parameter-efficient adapter with a cross-attention architecture. Text embeddings $E_t$ are used as learnable queries to attend to high-dimensional visual features $F_v$, extracted from intermediate layers of a CLIP vision encoder. This design ensures efficient distillation of identity-critical features, with minimal overhead relative to the base model. To enhance flexibility, DSM applies modulation hierarchically: a global offset adjusts the main textual condition, while finer-grained offsets are injected into individual layers of the DiT for dynamic identity reinforcement over the denoising process.

\section{Experiments}
\label{sec:experiments}

\subsection{Experiments Setting}

\paragraph{Implementation Details.}
We build our framework upon the pre-trained FLUX.1 dev \cite{flux2024} model, with a learning rate of $10^{-5}$ and a batch size of 1. Using a single-stage training strategy, we employ 32,000 high-quality text-pose pairs (128,000 images) and train the model for 100,000 steps. The experiments are conducted on 8 NVIDIA H200 GPUs, with LoRA \cite{hu2022lora} of rank 512 applied throughout training.

\paragraph{Comparative Methods.}
ASTRA is capable of handling both single-subject and multi-subject driven generation. For general subject-driven tasks, we compare it with several leading methods, including OmniGen \cite{xiao2025omnigen}, Ominicontrol \cite{tan2024ominicontrol}, FLUX IPAdapter v2 \cite{flux2024}, Ms-diffusion \cite{wang2024ms}, MIP-Adapter \cite{huang2025resolving}, RealCustom++ \cite{mao2024realcustom++},  SSR-Encoder \cite{zhang2024ssr} and UNO\cite{wu2025less}. For the more demanding task of multi-subject generation with complex pose control, where many of the aforementioned methods are not applicable, we benchmark against recent state-of-the-art models: OmniGen\cite{xiao2025omnigen}, DreamO\cite{mou2025dreamo} ,OmniGen2 \cite{wu2025omnigen2}, and UNO \cite{wu2025less}.

\paragraph{Evaluation Metrics.}
We adopt standard automatic metrics for our evaluation. Subject similarity is measured by CLIP-I~\cite{radford2021learning} and DINO~\cite{oquab2023dinov2} scores, while text fidelity is assessed with CLIP-T. Crucially, to quantify structural adherence, we employ Object Keypoint Similarity (OKS)~\cite{lin2014microsoft}. For subject-driven generation, our evaluation covers both single and multi-subject tasks. We assess single-subject performance on the standard DreamBench~\cite{ruiz2023dreambooth}. For multi-subject scenarios, following established protocols~\cite{huang2025resolving, shi2024instantbooth}, we use 30 subject combinations from DreamBench with 25 text prompts each, resulting in a comprehensive test set of 4,500 image groups.

To evaluate our core contribution in multi-subject complex pose control, we introduce a new challenging benchmark curated from the COCO Keypoints dataset~\cite{lin2014microsoft}, containing 1,000 images with up to three subjects. For each image, every annotated person is treated as a distinct subject: we use the COCO person bounding boxes to crop identity reference images, and construct a multi-person pose map by rasterizing all ground-truth keypoints into an OpenPose-style skeleton image. This skeleton map is then used as the pose condition for all pose-aware methods on this benchmark, while the same identity crops and text prompts are shared across methods. Concretely, for methods that officially support OpenPose-style or structural control inputs, we follow their recommended usage and feed our skeleton map through the corresponding pose-control branch. For methods that do not officially provide a dedicated pose interface, we implement our own support. In this way, every compared model receives an equivalent pose prior, ensuring that differences in performance reflect how identity and pose are fused rather than whether a method has access to pose information. 

\subsection{Qualitative Analyses}

We provide qualitative comparisons against state-of-the-art methods to visually verify the effectiveness of ASTRA in Figure~\ref{figure4}.

\subsection{Quantitative Evaluations}

\subsubsection{Automatic Scores.}
Table~\ref{tab:single_subject_adjusted} compares our methods on DreamBench \cite{ruiz2023dreambooth} against both tuning-based and tuning-free methods. As a tuning-free method, ASTRA has a leading scores over previous methods with the highest CLIP-I and CLIP-T scores of 0.847 and 0.330, and a competitive DINO score of 0.699. We also compare our method in the multi-image condition scenario in Table~\ref{tab:multi_subject_adjusted}. ASTRA achieves the CLIP-I and DINO scores of 0.745 and 0.554, and has competitive CLIP-T scores compared to existing leading methods. These show that ASTRA has a robust ability to preserve identity while adhering to text descriptions in both single and multi-subject contexts.

To evaluate the core challenge of multi-subject complex pose generation, we present the results on our benchmark derived from COCO Keypoints in Table~\ref{tab:coco_benchmark}. ASTRA demonstrates a significant lead in OKS score of 0.0452, confirming its superior ability to adhere to precise structural guidance. Critically, this top-tier pose accuracy is achieved without compromising identity, as evidenced by our state-of-the-art CLIP-I and CLIP-T scores of 0.7087 and 0.3194, and competitive DINO scores. This quantitatively validates that our framework effectively decouples appearance and structure, overcoming the trade-off that plagues existing methods.

\subsection{Ablation Study}
We conducted an ablation study on our COCO-based benchmark to validate the effectiveness of each key component in ASTRA. The results are shown in Table~\ref{tab:ablation}.

\paragraph{Effect of RAG-Pose.}
As shown in row (a), removing the RAG-Pose module causes a severe degradation in pose accuracy, with the OKS score plummeting from 0.0452 to 0.0095. This result demonstrates that RAG-Pose is essential for providing the clean and explicit structural priors required for high-fidelity pose control.

\paragraph{Effect of Asymmetric Positional Encoding.}
Rows (b) and (c) demonstrate the critical impact of our asymmetric design.Applying a uniform RoPE to all concatenated tokens (b) leads to a complete performance collapse in both identity (CLIP-I: 0.7087 → 0.3813) and pose (OKS: 0.0452 → 0.0041), proving it is incompatible with our direct token concatenation. While a uniform UNOPE (c) also fails, it particularly struggles to learn the pose (OKS drops to 0.0076), with other metrics degrading due to this inappropriate encoding. 

\paragraph{Effect of Disentangled Semantic Modulation (DSM).}
Removing the DSM stream (row d) results in a clear drop in identity preservation (CLIP-I: 0.7087 → 0.6754) while pose accuracy remains largely stable. This indicates that DSM successfully offloads the identity-learning task to a dedicated module, leading to more robust identity generation without interfering with the model's primary structural control task.

\begin{table}[h]
  \centering
  \small  
  \setlength{	\tabcolsep}{6pt}
  \begin{tabular}{lcccc}  
   \toprule
    \textbf{Method}  & \textbf{CLIP-I↑} & \textbf{DINO↑} & \textbf{CLIP-T↑} \\
    \midrule
    Oracle (reference images) & 0.885 & 0.774 & N/A \\
    \midrule
    Textual Inversion\cite{gal2022image}  & 0.780 & 0.569 & 0.255 \\
    DreamBooth\cite{ruiz2023dreambooth}  & 0.803 & 0.668 & 0.305 \\
    BLIP-Diffusion\cite{li2023blip}  & 0.805 & 0.670 & 0.302 \\
    \midrule
    ELITE\cite{wei2023elite}  & 0.772 & 0.647 & 0.296 \\
    Re-Imagen \cite{chen2022re} & 0.740 & 0.600 & 0.270 \\
    BootPIG \cite{purushwalkam2024bootpig}  & 0.797 & 0.674 & 0.311 \\
    SSR-Encoder\cite{zhang2024ssr} & 0.821 & 0.612 & 0.308 \\
    RealCustom++ \cite{mao2024realcustom++}  & 0.794 & \underline{0.702} & \underline{0.318} \\
    OmniGen\cite{xiao2025omnigen}  & 0.801 & 0.693 & 0.315 \\
    OminiControl\cite{tan2024ominicontrol}  & 0.799 & 0.684 & 0.312 \\
    FLUX.1 IP-Adapter\cite{flux-ipa} & 0.820 & 0.582 & 0.288 \\
    UNO\cite{wu2025less}   & \underline{0.835} & \textbf{0.760} & 0.304 \\
    \textbf{ASTRA(Ours)}  & 	\textbf{0.847} & 	0.699 & 	\textbf{0.330} \\
    \bottomrule
  \end{tabular}
  \caption{Quantitative results for single-subject driven generation on DreamBench. We present the oracle results in the first row and compare both tuning and tuning-free methods. The \textbf{best} results are shown in bold and the \underline{second-best} result is underlined.}
  \label{tab:single_subject_adjusted}
  \setlength{	\tabcolsep}{6pt}  
  
\end{table}

\vspace{-20pt}

\begin{table}[h]
  \centering
  \small  
  \setlength{	\tabcolsep}{7pt}  
  \begin{tabular}{lcccc}
   	\toprule
    	\textbf{Method}  & 	\textbf{CLIP-I↑} & 	\textbf{DINO↑} & 	\textbf{CLIP-T↑} \\
    \midrule
    DreamBooth \cite{ruiz2023dreambooth} & 0.695 & 0.430 & 0.308 \\
    BLIP-Diffusion \cite{li2023blip} & 0.698 & 0.464 & 0.300 \\
    \midrule
    SubjectDiffusion \cite{ma2024subject} & 0.696 & 0.506 & 0.310 \\
    MIP-Adapter \cite{huang2025resolving}  & 0.726 & 0.482 & 0.311 \\
    MS-Diffusion \cite{wang2024ms} & 0.726 & 0.525 & 0.319 \\
    OmniGen \cite{xiao2025omnigen}  & 0.722 & 0.511 & \textbf{0.331} \\
    UNO \cite{wu2025less} & \underline{0.733} & \underline{0.542} & 0.322 \\
    	\textbf{ASTRA (Ours)}  & 	\textbf{0.745} & 	\textbf{0.554} & 	\underline{0.326} \\
    \bottomrule
  \end{tabular}
  \caption{Quantitative results for multi-subject driven generation based on DreamBench. Our method achieves state-of-the-art performance among both tuning and tuning-free methods.}
 \label{tab:multi_subject_adjusted}
  \setlength{	\tabcolsep}{6pt}  
\end{table}
\vspace{-10pt}

\vspace{-5pt}
\begin{table}[h]
\centering
\small
\setlength{\tabcolsep}{4.5pt}
\begin{tabular}{lcccc}
\toprule
\textbf{Method} & \textbf{OKS} $\uparrow$ & \textbf{CLIP-I} $\uparrow$ & \textbf{DINO} $\uparrow$ & \textbf{CLIP-T} $\uparrow$ \\
\midrule
DreamO\cite{mou2025dreamo} & 0.0015 & 0.6736 & \underline{0.4665} & \underline{0.3084} \\
OmniGen\cite{xiao2025omnigen} & 0.0219 & 0.6865 & 0.4284 & 0.2999 \\
OmniGen2\cite{wu2025omnigen2} & 0.0270 & 0.6938 & \textbf{0.4670} & 0.3075 \\
UNO\cite{wu2025less} & 0.0277 & 0.6857 & 0.4283 & 0.2970 \\
IP-Adapter*\cite{flux-ipa} & \underline{0.0314}\ & \underline{0.6942}\ & 0.4310\ & 0.3050\ \\
PhotoMaker*\cite{li2024photomaker} & 0.0308\ & 0.6920\ & 0.4305\ & 0.3042\ \\
\textbf{ASTRA (Ours)} & \textbf{0.0452} & \textbf{0.7087} & 0.4457 & \textbf{0.3194} \\
\bottomrule
\end{tabular}
\caption{Evaluation on the COCO-based Complex Pose Benchmark. All compared methods receive identical multi-person pose maps and identity crops. Methods marked with * use \textbf{ControlNet and OpenPose} for explicit pose conditioning. ASTRA achieves the highest OKS while maintaining strong identity fidelity.}
\label{tab:coco_benchmark}
\setlength{\tabcolsep}{6pt}
\end{table}
\vspace{-10pt}

\begin{table}[h]
    \centering
    \small
    \setlength{	\tabcolsep}{1.5pt}
    \begin{tabular}{lcccc}
        	\toprule
        	\textbf{Configuration} & 	\textbf{OKS} $\uparrow$ & 	\textbf{CLIP-I} $\uparrow$ & 	\textbf{DINO} $\uparrow$ & 	\textbf{CLIP-T} $\uparrow$ \\
        \midrule
        (a) w/o RAG-Pose & 0.0095 & 0.6815 & 0.4173 & 0.3061 \\
        (b) w/ Symmetric RoPE & 0.0041 & 0.3813 & 0.2255 & 0.1705 \\
        (c) w/ Symmetric UNOPE & 0.0076 & 0.6028 & 0.3591 & 0.2876 \\
        (d) w/o DSM & 0.0345 & 0.6754 & 0.4208 & 0.3072 \\
        \midrule
        (e) 	\textbf{ ASTRA (Ours)} & 	\textbf{0.0452} & 	\textbf{0.7087} & 	\textbf{0.4457} & 	\textbf{0.3194} \\
        \bottomrule
    \end{tabular}
    \caption{Ablation study of ASTRA's key components on the COCO-based complex pose benchmark. The performance of the full model (e) is the main result, while rows (a-d) show the impact of removing or altering each component.}
    \setlength{	\tabcolsep}{6pt}
    \label{tab:ablation}
\end{table}

\section{Conclusion}
\label{sec:conclusion}
In this work, we address the challenge of generating multi-subject images with both high identity fidelity and precise pose control by identifying and resolving a core “positional encoding conflict” in existing models. We introduce ASTRA, a novel framework that architecturally disentangles appearance and structure by integrating three key components: a Retrieval-Augmented Pose (RAG-Pose) pipeline for clean structural priors, an asymmetric positional encoding strategy to separate identity and pose signals, and a Disentangled Semantic Modulation (DSM) stream to enhance identity robustness. Extensive experiments and ablation studies validate that our method effectively decouples the representation of “who” from “where,” achieving state-of-the-art results on complex pose and multi-subject benchmarks and paving the way for more controllable and reliable generative models. Beyond the benchmark tasks, the proposed design principles can be readily extended to other controllable generation scenarios involving structured or multi-modal conditions. As an additional contribution, we will release the curated pose database and the retrieval library used in our framework, providing valuable resources for future research in pose-conditioned and identity-preserving generation.
﻿

{
    \small
    \bibliographystyle{ieeenat_fullname}
    \bibliography{main}
}

% WARNING: do not forget to delete the supplementary pages from your submission 
% \input{sec/X_suppl}

\end{document}